\documentclass{article}

\usepackage[final]{neurips_2022}

\usepackage[utf8]{inputenc} % allow utf-8 input
\usepackage[T1]{fontenc}    % use 8-bit T1 fonts
\usepackage{hyperref}       % hyperlinks
\usepackage{url}            % simple URL typesetting
\usepackage{booktabs}       % professional-quality tables
\usepackage{amsfonts}       % blackboard math symbols
\usepackage{nicefrac}       % compact symbols for 1/2, etc.
\usepackage{microtype}      % microtypography
\usepackage{xcolor}         % colors

\usepackage[pdftex]{graphicx}
\usepackage{natbib}

\title{Identifying Spurious Correlations and Correcting them with an Explanation-based Learning}

\author{%
  Misgina Tsighe Hagos\\
  Science Foundation Ireland Centre for\\ Research Training in Machine Learning\\
  School of Computer Science\\
  University College Dublin\\
  Dublin, Ireland\\
  \texttt{misgina.hagos@ucdconnect.ie} \\
  \And
  Kathleen M. Curran\\
  Science Foundation Ireland Centre for\\ Research Training in Machine Learning\\
  School of Medicine\\
  University College Dublin\\
  Dublin, Ireland\\
  \texttt{kathleen.curran@ucd.ie} \\
  \AND
  Brian Mac Namee\\
  Science Foundation Ireland Centre for\\ Research Training in Machine Learning\\
  School of Computer Science\\
  University College Dublin\\
  Dublin, Ireland\\
  \texttt{brian.macnamee@ucd.ie} \\
}

\begin{document}

\maketitle

\begin{abstract}
  Identifying spurious correlations learned by a trained model is at the core of refining a trained model and building a trustworthy model. We present a simple method to identify spurious correlations that have been learned by a model trained for image classification problems. We apply image-level perturbations and monitor changes in certainties of predictions made using the trained model. We demonstrate this approach using an image classification dataset that contains images with synthetically generated spurious regions and show that the trained model was overdependent on spurious regions. Moreover, we remove the learned spurious correlations with an explanation based learning approach.
  %gaining the trust of the end-users of that model
\end{abstract}

\section{Introduction}
\label{introduction}

Most of the literature on explainable AI focuses on explaining a model's capabilities. For example, for image classification tasks, several approaches to feature attribution based model explanation generation are available. These approaches locate discriminative image regions used by models to classify categories \cite{zhou2016learning,selvaraju2017grad,omeiza2019smooth,wang2020score}. Such approaches, and other explainability methods, have found applications in several areas such as medical imaging \cite{belton2021optimising,hagos2022interpretable} and finance \cite{bussmann2020explainable,nicholls2021financial}. While explaining a model's predictions helps end users gain insights into how a model works, it is also important to explain the limits of a model to users to help them to understand when they should avoid using a model. Understanding and identifying the limits of a model is  equally as important as understanding the reasons for the outputs of a model.

%If we are able to locate parts of input image that a model finds salient, doesn't that mean the rest of the image is unimportant for the model--meaning can't we explain a model's limits by negating feature attribution explanations, one can argue. We believe a model's knowledge limits can't be presented with existing feature attribution based explanations for two reasons: (1) salient and non-salient image regions of a model are not binary, and (2) errors or model limits can only be explained with respect to the opposite class. 

\begin{figure}
   \centering
   \includegraphics[width=0.8\linewidth]{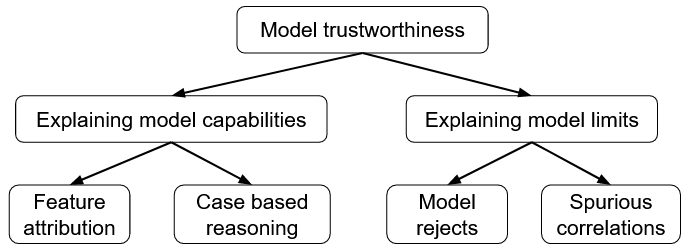}
   \caption{Model capabilities and limits overview.}
   \label{figure_capabilities_limits}
\end{figure}

%Due to the increased call for model explanations, specifying requirements that a proposed xAI method must meet helps for the design of better explanations. 
In 2020, The National Institute of Standards and Technology (NIST), an agency of the United States of America (USA) department of commerce, published "four principles of xAI" to which black-box models should adhere \cite{phillips2020four}. One of the four principles is \emph{knowledge limits}, which states that an AI system should explicitly present its shortcomings, and identify scenarios where it might fail. For these reasons, as is presented in Figure \ref{figure_capabilities_limits}, we view the process of achieving model trustworthiness through both understanding the capabilities of a model, and identifying and reducing model limits \footnote{Different terms such as \emph{knowledge limits} \cite{phillips2020four}, \emph{systemic errors} \cite{d2022spotlight}, \emph{failure analysis} \cite{singla2021understanding}, and \emph{blind-spot discovery} \cite{plumb2022evaluating} are used in the literature, but we believe \emph{model limits} generalizes all of them.}.

% \begin{enumerate}
%     \item \textbf{Explanation:} which states that an Artificial Intelligence's (AI) process or decision must be accompanied by reasoning.
%     \item \textbf{Meaningful:} which states that the explanation provided by an AI should be comprehensible for target users.
%     \item \textbf{Explanation Accuracy:} an AI's process or decision explanation must accurately describe the reasoning used by the AI.
%     \item \textbf{Knowledge Limits:} which states that an AI system should explicitly present its shortcomings and scenarios where it might fail.
% \end{enumerate}

\subsection{Model capabilities}

Considering whether model explanations utilize features or instances, we can categorize explanations of a model's capabilities into: (1) feature attribution, and (2) case based or example based explanations. Approaches to generate explanations based on feature attribution explain the predictions made by a model by identifying instance or image regions that the model finds salient. Class Activation Mapping (CAM) \cite{zhou2016learning} and Gradient Weighted Class Activation Mapping (GradCAM) \cite{selvaraju2017grad} are well known examples feature attribution approaches to explanation generation. Approaches to genrating expantions that are case based use other instances in order to explain the behaviour of a model. Counterfactual \cite{keane2020good} and semi-factual explanations \cite{kenny2021generating} are well known examples in this category. 

\subsection{Model limits}
\label{section_intro_model_limits}

In addition to explaining \emph{why} a model is working, it is equally important to explain model limits in which a trained model might not work. Model limits of a trained model are a set of instances, instance features or image regions that the model would fail to identify. They can be identified as instances that are out-of-distribution of a trained model \cite{plumb2022evaluating} or they can be located as spurious correlations or patterns a model might have picked from a training dataset \cite{singla2021salient}.

% \subsection{Spurious patterns}
% \label{spurious_regions}

Spurious patterns are parts of input instances that a model might have found salient during training, but that are considered unnecessary for classifying instances, or which are not causally related to the instance's category \cite{singla2021understanding}. For example, in the famous \emph{husky} vs \emph{wolf} classification task  popularized by Ribeiro et al. (2016) \cite{ribeiro2016should}, spurious regions that a model learned were identified from incorrectly classified pictures of a husky and a wolf \footnote{In the \emph{husky} vs \emph{wolf} classification, a model was trained with images of wolves that had snow in the background and of huskies that didn't. For testing the model, among other test instances, images of a husky that had a snow background and a wolf that didn't were used. This tricked the trained model into incorrect classification \cite{ribeiro2016should}.}. By creating test instances to trick the model, they showed that the model had learned spurious correlations using the confounding image regions that frequently appeared in the training instances and not the main objects in the test images. However, transferring this approach of identifying spurious correlations to other classification tasks would be expensive as it requires heavy manual input in identifying the candidate images that supposedly contain spurious regions.
%another example on spurious regions: horse vs car classifier where the horse images has tags. https://www.nature.com/articles/s41467-019-08987-4.pdf

%clever hans behaviour might also be worth mentioning

%Even though trained models can work correctly as expected during model testing phase, we still need to know if they will have limitations in the future. We define the model limits of a trained model as set of images, instance scenarios, or image regions that the model will fail to identify in the future, and that which we can not identify with certainty using existing feature attribution explanation methods.

In this paper, we present a simple perturbation based method to automatically identify spurious correlations learned by an image classifier, and an approach to unlearning them using explanation based learning. We use mean Monte Carlo (MC) dropout values to monitor model uncertainty and compare average changes in mean MC dropout values of perturbed images between a model trained with classification loss only and a model fine-tuned using explanation based learning. The following are the main contributions of this paper:

\begin{enumerate}
    \item We present an image-level perturbation based approach to identifying spurious correlations.
    \item We show how an image classifier model could be overly dependent on spurious image regions.
    \item We propose an explanation based learning approach to removing spurious correlations.
\end{enumerate}

\section{Related work}
\label{section_related_work}

% discuss why saliency maps are actually good to the contrary of 'sanity checks' paper, citing two other papers that refuted it??

The limits of a model can be presented at two levels. At the instance or image level, instances that are out of distribution for a model can be identified. This is usually implemented by adding \emph{reject options} to a model \cite{linusson2018classification}. With a reject option, a trained model is able to avoid making predictions for instances for which it has low certainty. At the feature level, spurious correlations, or confounding image features or regions can be identified. In this section, we present examples of both approaches to explaining the limits of a model and methods of unlearning them.

\subsection{Image level: out of distribution}

Image level explanation of model limits is concerned with identifying out of distribution instances and systemic errors \cite{plumb2022evaluating,singla2021understanding,d2022spotlight}. One type of case-based reasoning that were recently used for model limit explanation are semi-factual explanations \cite{kenny2021generating}. Artlet et al. (2022) \cite{artelt2022even} proposed to explain model rejects, which are out of distribution instances, with semifactual explanations that were generated using a perturbation based approach. However, because of the unlimited perturbation approach used, the explanations cannot be guaranteed to be plausible \cite{artelt2022even}. Online interviews were conducted to explain a model's error predictions in \cite{hlosta2020explaining}.

%TODO: explain why we also need feature level model limits?

%While image level identification of knowledge limits is helpful in listing cases where a model might fail and this informs users of when to avoid using the model, a deeper or feature level explanation of spurious correlations is important to understand the reason behind a model's limits and towards refining or retraining models to avoid spurious correlations. %This process of refining a model by interacting with its explanations so as to reduce spurious patterns of the model is termed as Explanatory Interactive Learning \cite{kulesza2015principles,teso2019explanatory} and is out of scope of this paper.

\subsection{Feature level: spurious correlations}

As is explained in Section \ref{section_intro_model_limits}, feature level identification of model limits involves locating or identifying spurious patterns. %This is best explained in Section \ref{section_intro_model_limits} with the \emph{husky} vs \emph{wolf} classifier \cite{ribeiro2016should}.

%Spurious patterns or spurious regions are image regions a model might have picked during training which are not parts of the object(s) of interest in an image or which are not causally related to the object's category \cite{singla2021understanding}. 

%To the best of our knowledge, Hlosta et al. 2020 \cite{hlosta2020explaining} and Artlet and Hammer 2022 \cite{artelt2022even}  are the only works that directly addressed the task of explaining a model's limits. 

% user input to annotate suspicious regions and examine if they are spurious patterns of the model \cite{plumb2021finding}

Most of the current literature on identifying spurious correlations learned by trained models requires an annotated dataset \cite{plumb2021finding}. Plumb et al. (2021) \cite{plumb2021finding} used pixel wise annotations available on the MS COCO dataset. Singla and Feizi et al. (2022) \cite{singla2021salient} use user annotations to identify spurious patterns of a model trained on the Imagenet dataset. But, involving users for pixel wise annotation of image datasets can be time-consuming and expensive. Singla et al. (2021) \cite{singla2021understanding} used image features extracted using a post-hoc explanation approach---Class Activation Mapping (CAM) saliency maps \cite{zhou2016learning}---and images that a trained model failed to correctly predict in order to associate image regions with errors. However, post-hoc explanations were reported to fail in identifying previously unknown spurious signals \cite{adebayo2021post}. The rate at which the predictions made by a model change after objects are added to or removed from input images  was also proposed as a way to identify spurious patterns \cite{plumb2021finding}. However, a model's prediction or probability outputs at the softmax or last layer of a model often results in an overconfident classification \cite{melotti2020probabilistic}.

% Plumb et al. (2022) showed that replacing image regions that a saliency maps explanation didn't show as salient in the first place could also lead to prediction change. 
%Even though the reported number of user annotation was minimal, the method in \cite{singla2021salient} still requires user input.

We identify expensiveness of involving users for image features annotation, failure of post-hoc explanations in identifying unknown spurious signals \cite{adebayo2021post} as research gaps in the current spurious correlations identification literature. Towards resolving both research gaps, we aim to automatically identify spurious correlations of a trained deep learning model without the need for user annotation or synthetically generated test instances.

%In addition to performing a search for spurious patterns on the usual candidates that are easily spotted with users, our method will cover image regions that are non-visible, such as brightness difference, as well.

Perturbation based explanations modify parts of an input instance and monitor changes in model output to identify regions that are important for model inference. LIME uses perturbation to generate subsamples for a simpler model training \cite{ribeiro2016should}. For image inputs, it identifies super-pixels and replaces them with mean values of corresponding regions \cite{garreau2021does}. Image perturbation can also be performed by sliding square boxes as masks over an input image \cite{zeiler2014visualizing,petsiuk2018rise}. However, perturbation can lead to implausible explanations \cite{artelt2022even}.

%To generate perturbed images that are in-distribution of the existing training and test dataset, we perform perturbation by exchanging image regions between a source image and subset target images of different category.

\subsection{Unlearning spurious correlations}
\label{related_work_unlearning}

Complying with regulations such as the \emph{right to be forgotten} of the GDPR \cite{mantelero2013eu,voigt2017eu} and removing polluted data \cite{liu2020learn} are some of the advantages of enabling a trained model to forget trained data or model unlearning. Model retraining and approximate learning are two of the major categories in model unlearning \cite{liu2020learn}. While model retraining removes effects of specific data points or instances from the training dataset by retraining a trained model without them \cite{neel2021descent,kim2022efficient}, approximate learning minimizes the computational cost of retraining a model by converting training algorithms into another form, a summation form for example \cite{cao2015towards}.

While model unlearning focuses on forgetting training data points at the instance level, a more detailed model refinement approach that utilizes instance or image features can provide a platform for rich communication with users which translates to better control over model refinement for users. Explanation based learning proposes to fill this gap by correcting models based on user interaction with- or feedback on model explanations \cite{stumpf2009interacting,kulesza2015principles}. Similar to model unlearning, there are two general categories of correcting trained models using explanation based learning: (1) model re-training, which involves generating new training examples to unlearn spurious correlations \cite{schramowski2020making,plumb2021finding}; and (2) model fine-tuning which adds an explanation loss into model training compared to its deviation or similarity from ground truth annotation of objects or spurious regions in an image, respectively \cite{ross2017right,teso2019explanatory,schramowski2020making,shao2021right}. %Instead of augmenting new training data, we propose to use existing annotations to fine-tune a model that was trained using classification loss only.

\section{Methods}
\label{section_methods}

%wolf vs husky exp srcs: https://www.kaggle.com/code/a45632/fastai-example-wolf-vs-husky
% https://www.kaggle.com/code/abhinavsp0730/why-i-trust-you

\subsection{Dataset}
\label{dataset}

To demonstrate the effectiveness of an explanation based learning to refine models, Teso and Kersting \cite{teso2019explanatory} used a decoy version of the Fashion MNIST dataset \cite{xiao2017fashion}. It is made up of Fashion MNIST images with 4x4 squares of pixels added to randomly selected image corners. However, the decoys used in this dataset are class independent, making it harder to track them and monitor their impacts on a model's certainty. We used a class-wise decoy version of the Fashion MNIST dataset that contains synthetic spurious regions in the same image regions across the same class \cite{hagos2022impact}. We chose this approach of adding class-wise spurious regions for two main reasons: (1) to monitor changes in model confidence after image region perturbations that include spurious regions using MC dropout; and (2) to take advantage of the available annotations of the spurious regions for an explanation based model fine-tuning. The dataset contains 60,000 28x28 pixel images for model training. A test dataset of 10,000 unseen images is used to measure model performance.

\subsection{Experimental setup}
\label{experimental_setup}

\subsubsection{Model training}
\label{model_training}

After experimenting with different architectures and comparing their performances, a convolutional neural network with one convolutional layer followed by a 50\% dropout and a fully connected layer was selected. We trained the model with the Adam optimizer and a decaying learning rate starting with \begin{math}1e^{-3}\end{math}. We used accuracy to measure classification performance, and the mean of 100 MC dropout forward passes to measure the trained model's certainty in classifying perturbed images. We use MC dropout to assess a trained model's certainty, to circumvent the overconfidence issue of softmax layers \cite{gal2016dropout}. We used Paperspace A4000 cloud service to train and test our models. It took us a total of 10 minutes to train a model using classification only, 42 minutes to train a model using an explanation based learning, and it took us an average of 17 minutes to compute mean MC dropouts of perturbed images using 50 images from a target class.

We ended up with two models that have the same architecture: an initial model that was trained using classification loss, \begin{math}L_{C} \end{math}, only (Equation \ref{eq:ross_classification_loss_2017}), and a fine-tuned version of the initial model using explanation based learning.  After we assess the initial model's confidence in classifying perturbed images and visualize the effects of spurious correlations, we fine-tune the model using existing spurious region annotations with an explanation based learning.

\begin{equation}
    \label{eq:ross_classification_loss_2017}
    L_{C} =\sum _{n=1}^{N}\sum _{k=1}^{K} -y_{nk}\log\hat{y}_{nk} 
\end{equation}

For the explanation based model fine-tuning, we used a Right for the Right Reasons (RRR) using GradCAM (Equation \ref{eq:schramowski}) explanation loss, \begin{math}L_{E} \end{math}, that is computed between GradCAM model explanations and spurious region annotations. L2 regularization is also used to address over-fitting. This is summarised in Equations \ref{eq:schramowski} and \ref{eq:class_exp_loss} using GradCAM explanations, where \begin{math}M_{n} \in \ \{0,\ 1\}\end{math} is the ground truth annotation and \begin{math} norm \end{math} normalizes the Grad-CAM output, \begin{math} \theta \end{math} holds a model's parameters, with input \begin{math} X \end{math}, labels \begin{math}y \end{math}, predictions \begin{math}\hat{y} \end{math}, and a parameter regularization term \begin{math} \lambda \end{math}.

\begin{equation}
    \label{eq:schramowski}
    L_{E} =\ \sum _{n=1}^{N}( M_{n} * \ norm( GradCAM_{\theta }( X_{n})))^{2}
\end{equation}

\begin{equation}
    \label{eq:class_exp_loss}
    Loss\ =\ L_{C} + L_{E} +\ \lambda\sum _{i} \theta _{i}^{2}
\end{equation}

\begin{figure}%[t]
  \centering
  \includegraphics[width=0.9\linewidth]{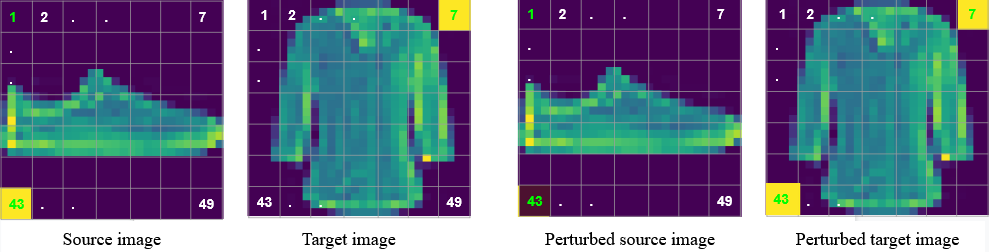}
  \caption{Example image perturbation between an image from source class, \emph{sneaker}, and target class, \emph{Shirt} (Left). Perturbations outputs, by exchanging image patches that are numbered 43, are displayed in the two images in the right. Numbered grids are added to show image regions in source image and their corresponding regions in the target image with which they were exchanged for perturbation.}
  \label{image_perturbation}
\end{figure}

\subsubsection{Perturbation}
\label{method_perturbation_based}

To perform perturbation we randomly select two classes: source class $c_{s}$, target class $c_{t}$, an input image $x_{s}$ categorized as $c_{s}$, and a set of images, $X_{t}$ (size=N) categorized as $c_{t}$. We apply a 4x4 square of pixels exchange, $R$, between $x_{s}$ and each image in $X_{t}$, $R(x_{s}, X_{t}) = X^{*}_{r}$. For the decoyed Fashion MNIST, we use images from the \emph{Sneaker} category as the source class and subset of images (N=50) from the \emph{Shirt} category as target class. For single images from the source and target classes, we end up with 49 perturbations given the 28x28 size of Fashion MNIST images. Two example images from the source and target classes are presented in Figure \ref{image_perturbation}, where the perturbations are applied by exchanging image patches between the corresponding numbers in the grid. We perform perturbation using images in the existing dataset to generate perturbed images that are in-distribution of the existing training and test dataset. This grounds the generated perturbations to the space of available dataset--making our results plausible.

\subsection{Identifying spurious correlations}
\label{method_monitor_certainty}

We associate a trained model's limits with its ability to correctly classify the perturbed versions of images. Conformity measure has been used to compute a model's certainty and to identify out of distribution instances \cite{artelt2022even}, but since the perturbed images we have are in-distribution we resort to the MC dropout.

After applying perturbation of a 4x4 square of pixels exchange, $R$, between $x_{s}$ and each image in $X_{t}$, $R(x_{s}, X_{t}) = X^{*}_{r}$, as is discussed in Section \ref{method_perturbation_based}, we then monitor the trained model's, $f$, certainty in classifying the perturbed images, using a mean MC dropout value of a 100 loops, $f_{MC}(X^{*}_{r})$. We locate image regions that were used for perturbations that brought the biggest negative impact on the model's certainty and resulted in the lowest certainty value: $argmin(f_{MC}(X^{*}_{r}))$. Given the over-dependence of the model, we take these image regions as candidates of spurious correlations and use an explanation based learning (Equation \ref{eq:class_exp_loss}) described in Section \ref{model_training} to fine-tune the trained model. We perform this fine-tuning to reduce effects of the spurious regions on the model's prediction.

%\subsubsection{Evaluation}

%Plausibility. This could be visual, by adding a few sample perturbed images and their corresponding saliency maps and arguing how realistic/ plausible they are. if there is time, uncertainty estimation of the model on the perturbed images could be computed. If uncertainty is low leading to reject, they are not plausible, but if it's high--realistic images.

    % \vspace{\floatsep}
\begin{figure}[t]
   \centering
   \includegraphics[width=0.9\linewidth]{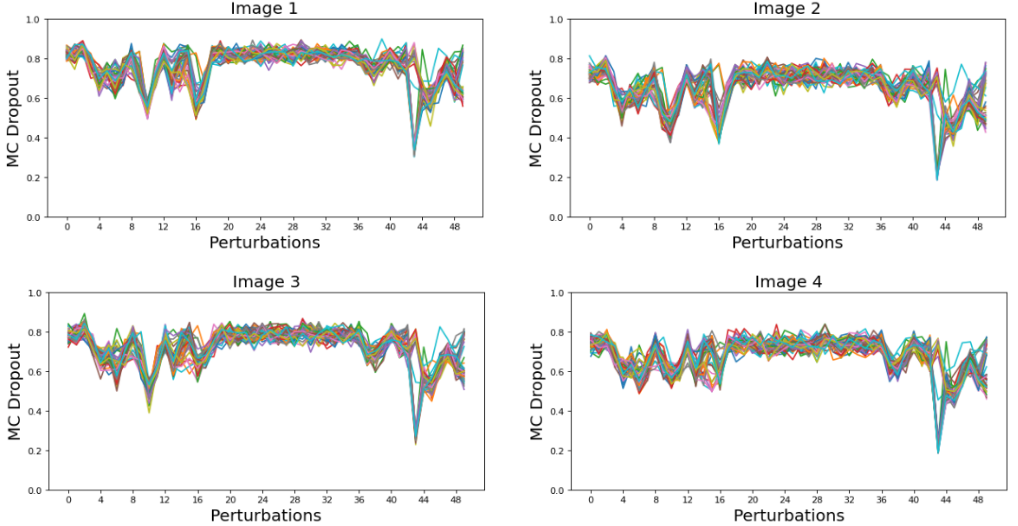}
   \caption{Mean MC Dropout values of perturbed images using a model trained with classification loss only.}
   \label{image_mc_dropout_unrefined}
    
    \vspace{\floatsep}
    
   \centering
   \includegraphics[width=0.9\linewidth]{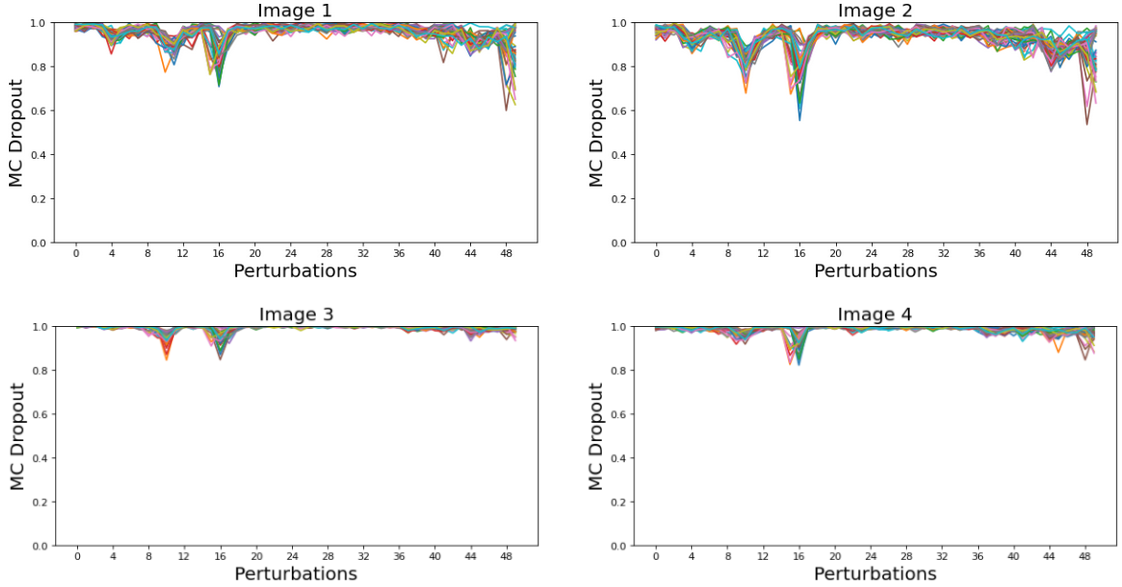}
   \caption{Mean MC Dropout values of perturbed images using a model that was refined using explanation based learning.}
   \label{image_mc_dropout_refined}
\end{figure}

\section{Results and conclusion}
\label{section_results}

Figure \ref{image_mc_dropout_unrefined} shows MC dropout values of four test images from class \emph{Sneaker} that were initially correctly classified by the model, after each perturbation against 50 images from the target class, class \emph{Shirt}. At perturbation number 43 (which perturbs the spurious region as seen in Figure \ref{image_perturbation}), we can see that the model's mean MC values suddenly drop for all the plots. We normally expect a change of mean MC values as a result of the perturbations, but the changes in mean MC values at the 43rd perturbation are visually much higher than other perturbations. This is because the spurious region in the test images were masked by zeroed regions from the target images. This shows that the trained model was overly dependent on the spurious regions for its prediction. Some of the perturbed images didn't lose much MC values because the target regions that were used for perturbation had higher pixel intensities and weren't zero.

After visualizing the effects of spurious correlations on mean MC dropout values, we refined the trained model using an explanation based learning (Equation \ref{eq:class_exp_loss}). Mean MC dropout values of the same test images used in Figure \ref{image_mc_dropout_unrefined} using the resulting refined model and using the same 50 images from the target class are seen in Figure \ref{image_mc_dropout_refined}. While there are still changes in MC values as a result of perturbations, the huge changes seen in Figure \ref{image_mc_dropout_unrefined} were removed with an explanation based fine-tuning. In addition to the visual conformation of the removal of spurious correlations, while the average change in mean MC values for the original model was at 0.50, the refined model only had an average change of MC values at 0.15.

% \section{Conclusion}
% \label{section_conclusion}

With a dataset that consisted of synthetically generated spurious regions and by monitoring mean MC dropout values on perturbed images, we were able to see that a trained model was overly dependent on spurious regions. This could be because the spurious regions are on the same location across images of same category in the training dataset. This approach of identifying spurious correlations can be extended to other image classification areas that could naturally have spurious regions in their training dataset. In addition to spotting spurious correlations, we were able to decrease their effects on a model's certainty with an explanation based learning. By explaining instance feature level limits of a trained model and providing end users with a way to correct spurious correlations, we believe our method can contribute towards developing trustworthy models.

Although we used the existing spurious regions annotations, as user feedback, to implement the explanation based learning, our approach can also be extended to build an explanation based interactive learning where users can actively participate in the model building process. In addition, in our experiments, while the artificially added spurious regions brought the highest negative impacts on a trained model's certainty, a wider experiment involving other datasets is required to assess if perturbing other image regions could have higher impacts than the spurious regions. For future work, we plan to expand our approach to other datasets that contain confounding regions and to involve user interactions as feedback on model explanations instead of using existing annotations.

\section*{Acknowledgements}

This publication has emanated from research conducted with the financial support of Science Foundation Ireland under Grant number 18/CRT/6183. For the purpose of Open Access, the author has applied a CC BY public copyright licence to any Author Accepted Manuscript version arising from this submission.

%If you cite your other papers that are not widely available(e.g., a journal paper under review), use anonymous author names in the citation, e.g., an author of the form ``A.\ Anonymous.''

%\section*{Acknowledgements}

\bibliographystyle{splncs04}
\bibliography{neurips_2022}

%%%%%%%%%%%%%%%%%%%%%%%%%%%%%%%%%%%%%%%%%%%%%%%%%%%%%%%%%%%%
\section*{Checklist}

\begin{enumerate}

\item For all authors...
\begin{enumerate}
  \item Do the main claims made in the abstract and introduction accurately reflect the paper's contributions and scope?
    \answerYes{}
  \item Did you describe the limitations of your work?
    \answerYes{}{}
  \item Did you discuss any potential negative societal impacts of your work?
    \answerNA{}
  \item Have you read the ethics review guidelines and ensured that your paper conforms to them?
    \answerYes{}
\end{enumerate}

\item If you are including theoretical results...
\begin{enumerate}
  \item Did you state the full set of assumptions of all theoretical results?
    \answerNA{}
        \item Did you include complete proofs of all theoretical results?
    \answerNA{}
\end{enumerate}

\item If you ran experiments...
\begin{enumerate}
  \item Did you include the code, data, and instructions needed to reproduce the main experimental results (either in the supplemental material or as a URL)?
    \answerNo{}{We have included the instructions to reproduce the experiments (See Section \ref{section_methods}). We will release the source code soon.}
  \item Did you specify all the training details (e.g., data splits, hyperparameters, how they were chosen)?
    \answerYes{}
        \item Did you report error bars (e.g., with respect to the random seed after running experiments multiple times)?
    \answerNo{}
        \item Did you include the total amount of compute and the type of resources used (e.g., type of GPUs, internal cluster, or cloud provider)?
    \answerYes{}
\end{enumerate}

\item If you are using existing assets (e.g., code, data, models) or curating/releasing new assets...
\begin{enumerate}
  \item If your work uses existing assets, did you cite the creators?
    \answerYes{}
  \item Did you mention the license of the assets?
    \answerNA{}
  \item Did you include any new assets either in the supplemental material or as a URL?
    \answerNA{}
  \item Did you discuss whether and how consent was obtained from people whose data you're using/curating?
    \answerNA{}
  \item Did you discuss whether the data you are using/curating contains personally identifiable information or offensive content?
    \answerNA{}
\end{enumerate}

\item If you used crowdsourcing or conducted research with human subjects...
\begin{enumerate}
  \item Did you include the full text of instructions given to participants and screenshots, if applicable?
    \answerNA{}
  \item Did you describe any potential participant risks, with links to Institutional Review Board (IRB) approvals, if applicable?
    \answerNA{}
  \item Did you include the estimated hourly wage paid to participants and the total amount spent on participant compensation?
    \answerNA{}
\end{enumerate}

\end{enumerate}

%%%%%%%%%%%%%%%%%%%%%%%%%%%%%%%%%%%%%%%%%%%%%%%%%%%%%%%%%%%%

\end{document}